\pdfoutput=1

\documentclass[letterpaper, 10 pt, conference]{ieeeconf}  

\IEEEoverridecommandlockouts                              

\usepackage[left=54pt,right=54pt,top=54pt,bottom=54pt]{geometry}

\usepackage{adjustbox}
\usepackage{graphicx}
\usepackage{amsmath}
\usepackage{amssymb}
\usepackage{color}
\usepackage{subfigure}
\usepackage{url}
\usepackage{placeins}
\usepackage{array}
\usepackage{tabularx}
    \newcolumntype{L}{>{\raggedright\arraybackslash}X}

\graphicspath{ {images/} }

\newlength{\imagewidth}

\title{\LARGE \bf
Crop Height and Plot Estimation for Phenotyping\\ from Unmanned Aerial Vehicles using 3D LiDAR
}

\author{Harnaik Dhami, Kevin Yu, Tianshu Xu, Qian Zhu, Kshitiz Dhakal, James Friel, Song Li, and Pratap Tokekar
\thanks{This material is based upon work supported in part by NIFA grants 2018-67007-28380 and 2018-51181-28384.}
\thanks{H. Dhami, T. Xu, and P. Tokekar were with the Department of Electrical and Computer Engineering,
        Virginia Tech, U.S.A. when part of the work was completed.
        They are currently with the Department of Computer Science,
        University of Maryland, U.S.A.
        {\tt\small \{dhami, txu, tokekar\}@umd.edu}}%
\thanks{K. Yu is with the Department of Electrical and Computer Engineering,
        Virginia Tech, U.S.A. 
        {\tt\small klyu@vt.edu}}
\thanks{Q. Zhu, K. Dhakal, J. Friel, and S. Li are with the School of Plant and Environmental Sciences,
        Virginia Tech, U.S.A.
        {\tt\small \{qianzi19, kshitiz, jamesfriel, songli\}@vt.edu}}%
}

\begin{document}

\maketitle
\thispagestyle{empty}
\pagestyle{empty}

\begin{abstract}

We present techniques to measure crop heights using a 3D Light Detection and Ranging (LiDAR) sensor mounted on an Unmanned Aerial Vehicle (UAV). Knowing the height of plants is crucial to monitor their overall health and growth cycles, especially for high-throughput plant phenotyping. We present a methodology for extracting plant heights from 3D LiDAR point clouds, specifically focusing on plot-based phenotyping environments. 
We also present a toolchain that can be used to create phenotyping farms for use in Gazebo simulations. The tool creates a randomized farm with realistic 3D plant and terrain models. 
We conducted a series of simulations and hardware experiments in controlled and natural settings. Our algorithm was able to estimate the plant heights in a field with 112 plots with a root mean square error (RMSE) of 6.1 cm. This is the first such dataset for 3D LiDAR from an airborne robot over a wheat field. The developed simulation toolchain, algorithmic implementation, and datasets can be found on our GitHub repository.\footnote{\url{https://github.com/hsd1121/PointCloudProcessing}}

\end{abstract}

\section{INTRODUCTION}

The goal of precision agriculture is to optimize the growth, maintenance, and harvesting of crops using data-driven technologies~\cite{rogers,tokekar2016sensor}. This will become especially important as the population grows, leading to a higher demand of efficiency from farms~\cite{bommarco_kleijn_potts_2013, franzluebbers_paine_winsten_krome_sanderson_ogles_thompson_2012, godfray_2011}. One way of achieving higher efficiency is through selective breeding with the aid of plant phenotyping. Phenotyping refers to collecting large scale phenotypic information about plants, including their heights, which are used for selective breeding and associative mapping through, for example, genome-wide association studies.

A bottleneck in phenotyping is the data collection process. One important phenotypic trait to monitor during a plant's growth cycle is its height. Recording the plant growth allows agronomists to monitor and predict vital features of crops such as flowering time and yield~\cite{moles_leishman, moles_warton_warman_swenson_laffan_zanne_pitman_hemmings_leishman_2009}. Manual height measurements are labor-intensive and quickly become infeasible as the plot and farm size increases. Manual height measurements are also sometimes biased. Using wheat as one example, it is only feasible to measure a few selected plants in each plot where hundreds of wheat plants are grown. In this paper, our goal is to alleviate this bottleneck by using a UAV equipped with a 3D LiDAR for plant height estimation.



\begin{figure}[htp]
\centering
\includegraphics[width=\linewidth]{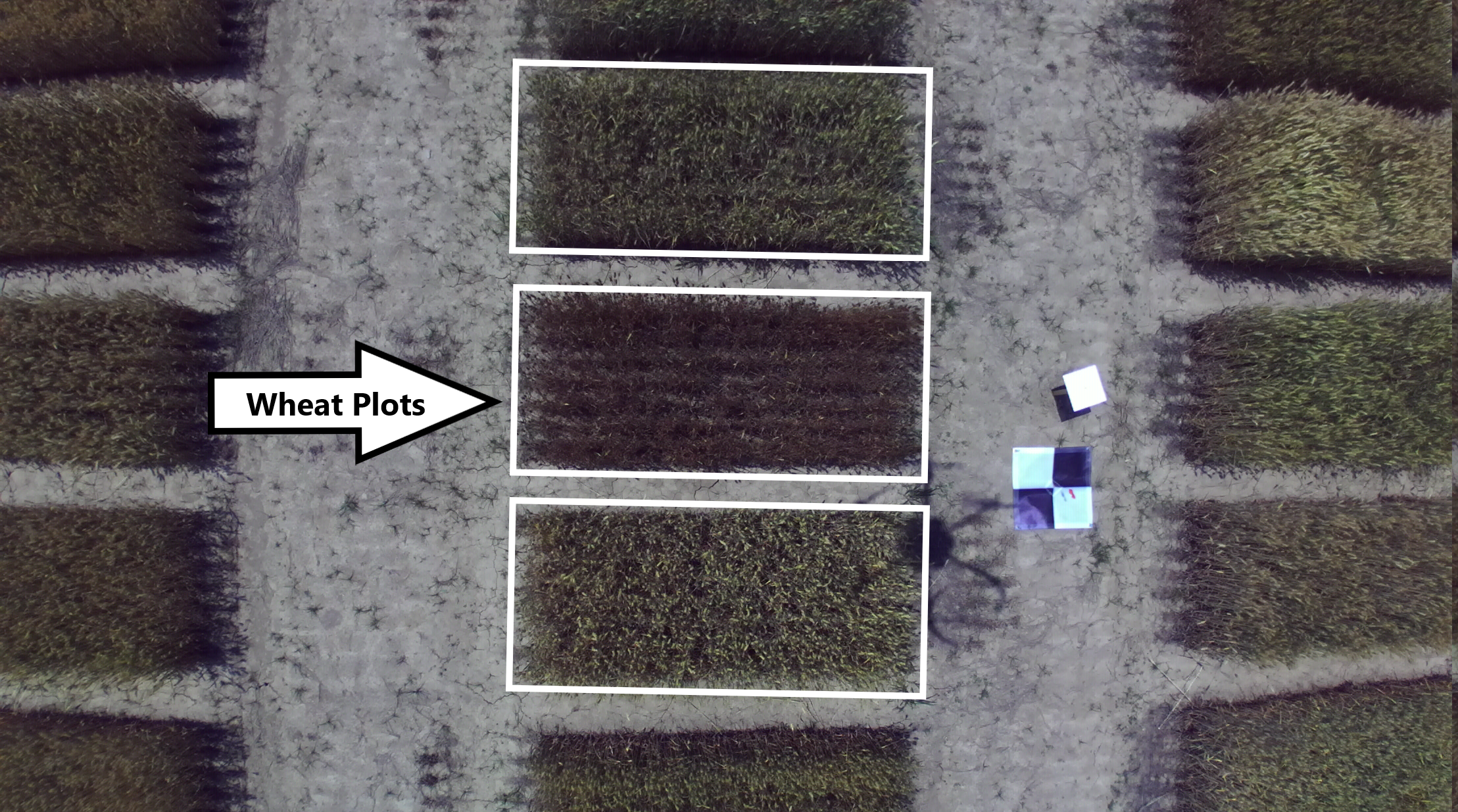}
\caption{This overhead picture was taken during a flight over wheat crops by our UAV. The farm is organized into plots, indicated by white boxes. The plots are organized as a grid. Our technique finds the plots and estimates the height of the crops within a plot.}
\label{fig:wheat_rgb}
\end{figure}

We present a technique for determining crop heights using a 3D LiDAR mounted on a UAV. In particular, we focus on farms organized into smaller plots as shown in Figure~\ref{fig:wheat_rgb}, as is typical in the case of phenotyping studies. Each plot (or a collection of plots) are a specific breed of plants being cultivated and monitored by the agronomist. The best cultivars are then selected for breeding. Traditionally, these plots are monitored by manual height measurements. Instead, we show how to use LiDAR to enable scalable and high-throughput phenotyping.

We collect raw 3D LiDAR scans from a UAV flown above the farm. We then present several data processing techniques that produce as output, bounding boxes around individual plots as well as height estimates for the plants within each plot. We do not make any assumptions on the prior knowledge about the size of each plot or the flatness of the terrain. We also present a toolchain to generate virtual phenotyping farms that can be used in simulations. In the real world, it is hard to find multiple farms to test the robustness of algorithms. This makes it difficult to compare algorithms or for robotics researchers to conduct precision agriculture research. Our goal in designing this toolchain is to make this research more accessible to the rest of the community. 

While there has been recent work on plant height estimation with LiDAR, as we describe in Section~\ref{chap:relWork}, this is the first such work using a 3D LiDAR mounted on a UAV that also detects individual plots. The two main contributions of this paper are as follows:
\begin{itemize}
  \item A toolchain to generate virtual plot-based farms with varying ground terrains useful for phenotyping. The dataset released along with this paper has models for three representative environments, simulated 3D LiDAR scans, and ground truth information. This is released for the community-at-large to benchmark their algorithms against.
  \item A suite of point cloud processing tools for 3D LiDAR scans collected from a UAV on a wheat farm. These include open-loop mapping, data-filtering, plot clustering, and the height estimation tool. The datasets and the point cloud processing tools are also released along with this paper. Our technique was able to estimate the plant heights of the real-life wheat plots with an RMSE of 6.1 cm.
\end{itemize}

We describe the hardware and software setup in Section~\ref{chap:systemDescription}. We describe our algorithm for plot and height estimation along with the underlying assumptions in Section~\ref{chap:rowAlgorithm}. The virtual farm generation toolchain is discussed in Section~\ref{chap:farmGen}. We conducted both hardware experiments and simulations with real wheat plots and virtual soybean plots. Lastly, we present the results and observations in Section~\ref{chap:experiments} followed by improvements we are working on in Section~\ref{chap:conclusionFutureWork}.\footnote{This paper is best viewed in color for better visualization of the figures.}





\section{Related Work}
\label{chap:relWork}

Automatically estimating the height of plants is an important problem and as such, there has been recent work on this (Table~\ref{table:comp}). However, as we will describe in this section, there are key differences between the prior work and the work presented in this paper.
Prior methods include using a 2D LiDAR mounted on a UAV~\cite{anthony_elbaum_lorenz_detweiler_2014}, using RGB cameras mounted on fixed-wing~\cite{ziliani_parkes_hoteit_mccabe_2018} and multi-rotor UAVs~\cite{madec, yuan_li_bhatta_shi_baenziger_ge_2018}, and using a ground robot for navigation between rows of crops~\cite{kayacan_young_peschel_chowdhary_2018}.

The method described by Anthony et al. uses a 2D LiDAR mounted on the bottom of a UAV facing downwards to measure corn heights~\cite{anthony_elbaum_lorenz_detweiler_2014}. They estimate the ground and the crown of the crop plants for each scan. They reported the accuracy of height estimations within 5 cm. However, they do not produce a 3D map of the farm or find individual plots, as we do in this paper.

Madec et al. used a similar method, except with a ground-based, 3D LiDAR on a row-based wheat farm along with a UAV equipped with an RGB camera~\cite{madec}. Using structure-from-motion, they extract a 3D dense point cloud from a camera on a UAV. 
They found a strong correlation between the estimated heights from the UAV and the ground-based LiDAR. The LiDAR estimations had an RMSE of 3.5 cm while the UAV estimations had an RMSE that ranged from 2.6 - 6.8 cm. It is expected for the ground-based method to be more accurate since a UGV does not have to constantly adjust its position to maintain stability. These adjustments add noise to the air-based measures. Unlike their method, we find each of the wheat plots as opposed to arbitrarily breaking up the 3D point cloud.

Yuan et al. used a similar setup with the 3D LiDAR mounted on a ground vehicle~\cite{yuan_li_bhatta_shi_baenziger_ge_2018}. 
The ground vehicle drove among the plots of the row-based wheat crops. To determine the height of the crops using the LiDAR, they relied on finding the ground in the areas between plots. The distance of the LiDAR off the ground was manually recorded and they used this distance to estimate the heights within the scans. Their ground-based LiDAR system had an RMSE of 5 cm; whereas the UAV approach had an RMSE of 9 cm. As opposed to manually determining the distance between the LiDAR and the ground, our method automates this process through the algorithm.

Ziliani et al. used a fixed-wing UAV with a downwards-facing camera ~\cite{ziliani_parkes_hoteit_mccabe_2018} to determine intra-field variability within a farm of maize crops. 
They used ground control points to help calibrate the image data and estimate the height. Also, to validate the estimations, they used a ground-based LiDAR mounted on a vehicle. There was a correlation of up to 0.99 between the LiDAR and structure-from-motion based approach for the RGB images. During the flowering of the crops, there was an increased variability and the correlation decreased to 0.65. However, they do not cover the entire farm with their 3D LiDAR as we do with our method.

There has also been some work done in regards to row detection by Li et al.~\cite{corn_row_detection:online}. They flew a UAV overhead a field of corn and captured RGB images from a camera. These RGB images were then stitched together to create a single image. They detect the rows of corn using computer vision techniques. Our method focuses on plot detection as opposed to row detection due to how our target farms are planted.

A detailed comparison between these prior works and ours is presented in Table~\ref{table:comp}. Our techniques can estimate the heights within an RMSE of 6.1 cm which is comparable to the other techniques. However, there are key differences between ours and the prior work: None of the previous works used a 3D LiDAR on the UAV to estimate the height. They either used a 2D LiDAR or RGB cameras. If a 3D LiDAR was used, it was only from the ground (and therefore restricted to either the edges of the farm or between the rows) and for validating the RGB data. Unlike prior work, we estimate the ground plane using point-cloud analysis. Anthony et al.~\cite{anthony_elbaum_lorenz_detweiler_2014} did this for individual 2D line scans whereas we do this for local 3D patches. We also detect bounding boxes around each plot in the 3D point cloud map of the entire farm. This is particularly useful during phenotyping. 


\begin{table}[htp]
\centering
\includegraphics[width=\linewidth]{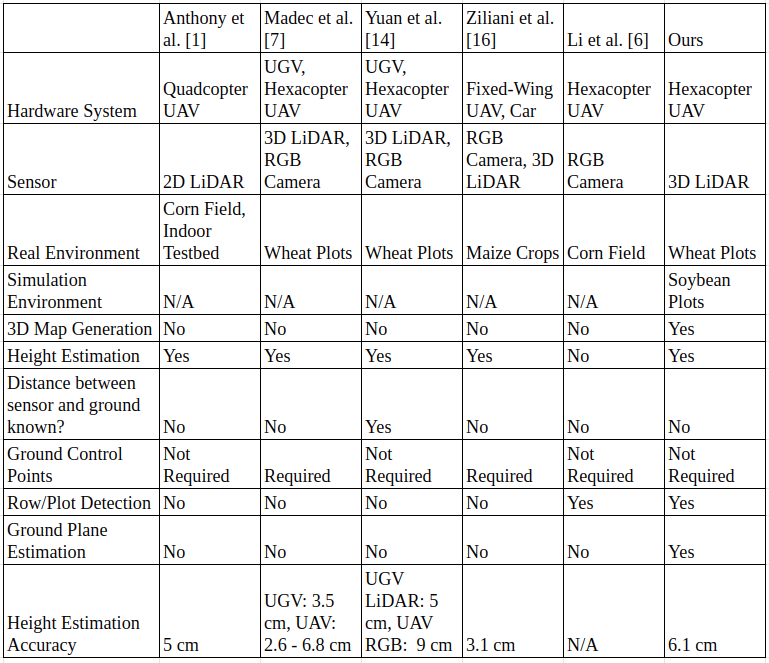}
\caption{Comparison between related work and our paper.}
\label{table:comp}
\end{table}




\section{System Description} \label{chap:systemDescription}



\begin{figure}[ht] 
  \subfigure[]{%
    \includegraphics[width=.22\textwidth]{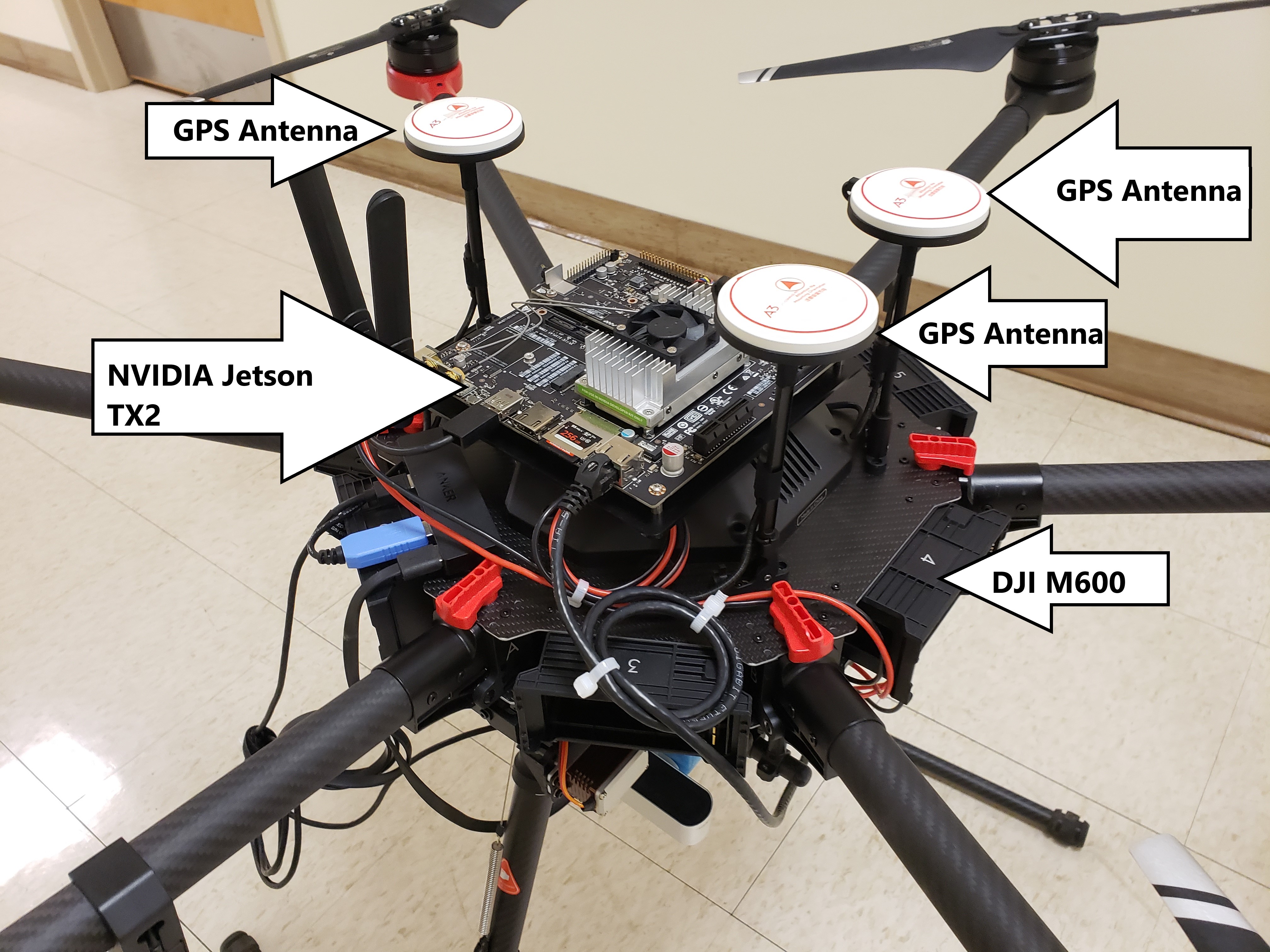} 
  } 
  \quad 
  \subfigure[]{%
    \includegraphics[width=.22\textwidth]{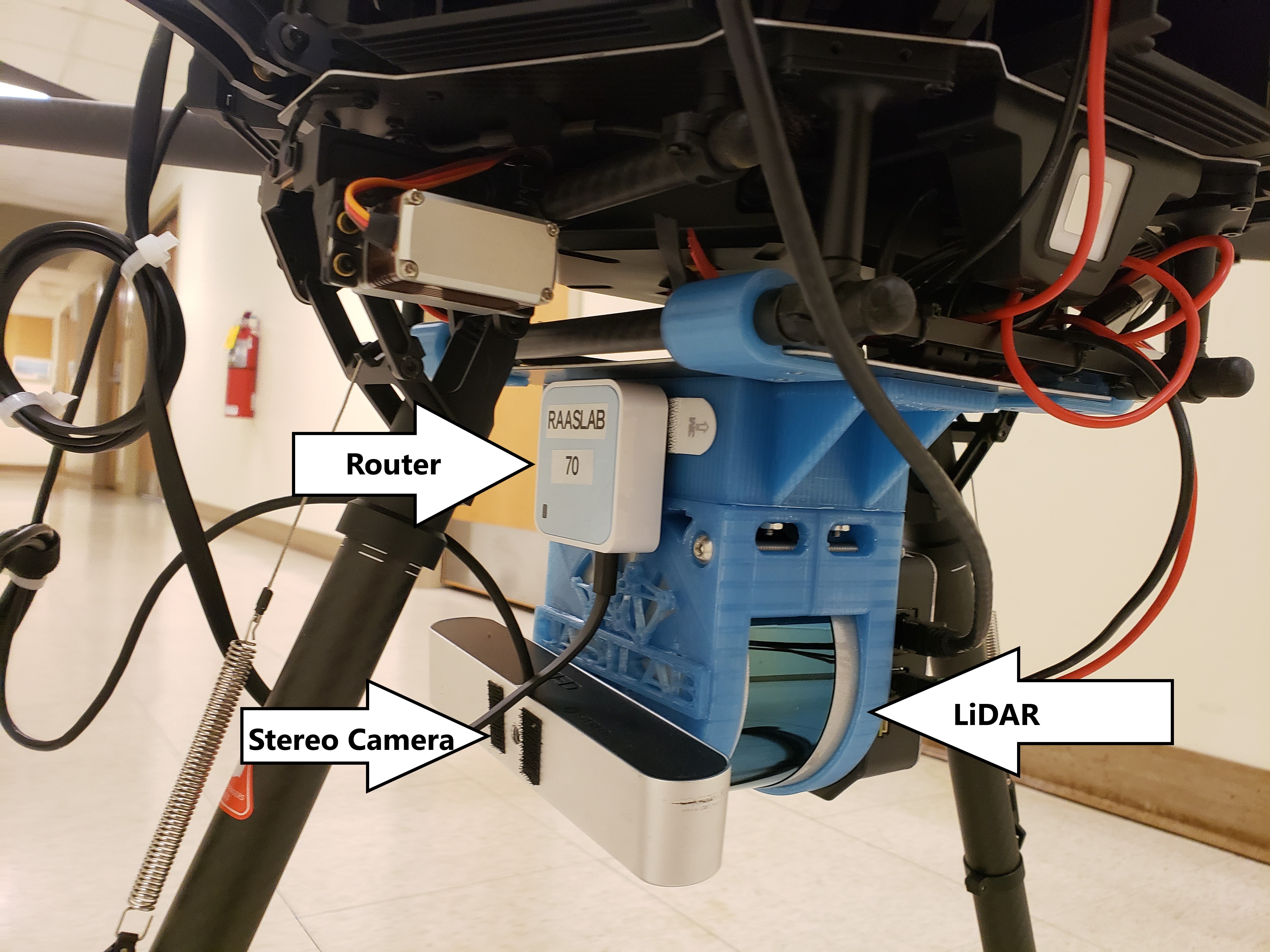}
  } 
  \caption{The selected hardware used during experimentation. (a) The UAV platform is a DJI M600 Pro that has 3 GPS receiver antennas. The onboard computer is an NVIDIA Jetson TX2. (b) Underneath, the Velodyne VLP-16 3D LiDAR is mounted.} 
  \label{fig:hardware_mounted}
\end{figure}

The UAV platform that we use is the DJI Matrice 600 Pro. The on-board computer is the NVIDIA Jetson TX2. The 3D LiDAR of choice is the Velodyne VLP 16. 
The DJI Matrice 600 pro has a maximum takeoff weight of 15.5 kg. The platform has a maximum speed of 17.88 m/s if there is no wind and a hovering time of 16 minutes with a 6 kg payload. The VLP-16 is a 3D LiDAR consisting of 16 channels that can refresh at a rate of 5--20 Hz. It also has a range of 100 meters and an accuracy of $\pm3$ cm. The 16 channels of the 3D LiDAR, compared to a single channel for a 2D LiDAR, allow for the use of more information during the crop height estimation pipeline. Figure~\ref{fig:hardware_mounted} shows all the hardware mounted on the DJI M600 Pro platform. 
We used NVIDIA Jetpack 3.2.1 on the TX2 with Ubuntu 16.04 installed, as well as NVIDIA's CUDA 9.0 and OpenCV. We also used Robot Operating System (ROS) Kinetic along with the DJI SDK and the Point Cloud Library~\cite{Rusu_ICRA2011_PCL} (PCL).

\section{Data Products: Height Estimation, Plot Detection, and 3D Maps} \label{chap:rowAlgorithm}

\begin{figure}[htp]
\centering
\includegraphics[width=\linewidth]{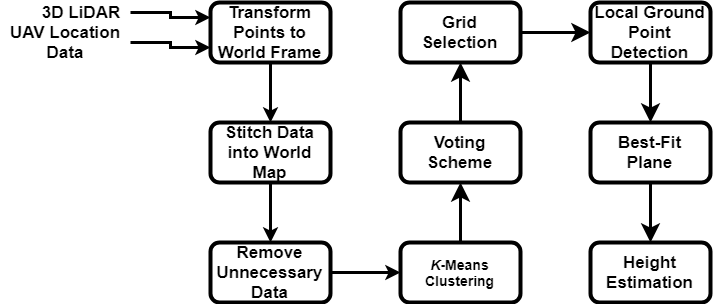}
\caption{The system block diagram describing our height estimation algorithm.}
\label{fig:block}
\end{figure}

Figure~\ref{fig:block} shows the system block diagram for the algorithm pipeline. We discuss each of the blocks step-by-step below.
We make the following assumptions based on the structure of typical phenotyping farms. We assume that the plots are aligned in a regular ordered grid locally. We also know the number of plots. In principle, we only need to know the number of plots for a small subset of the environment. These are justified in phenotyping studies since the agronomists are the ones who decide the layout of the farms. Nevertheless, what the agronomists need for scalable and high-throughput phenotyping are 3D maps of the farm, to automatically find bounding boxes around individual plots, and to find the height relative to the ground for each plot. We describe the point cloud processing tools that we have developed to generate these data products in this section.

\subsection{Pre-processing and Map Building}

We transform each incoming scan into the world coordinate frame and merge them to produce a 3D map. 
There are several open-source software packages that can be used such as LOAM~\cite{Zhang-2014-7903} and BLAM~\cite{nelson_2016}. However, we found that in our setting this was not necessary since the pose obtained by the UAV was good enough to generate an open-loop map. 
The DJI M600 fuses information from three GPS receivers along with the IMU data to obtain its global pose. Since we are operating on a farm, the UAV has a clear line-of-sight with the GPS satellites which, along with the three GPS corrections, yields satisfactory mapping. Nevertheless, any of the open-source mapping tools can be used.

\begin{figure}[htp]
\centering
\includegraphics[width=\linewidth]{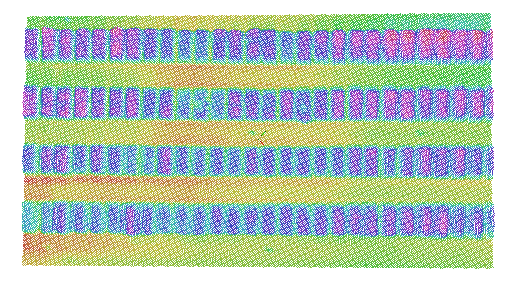}
\caption{This is an example output of the processing done on the point cloud scans. These scans were from the Kentland wheat farm. The colors represent the height of the points.}
\label{fig:wheat}
\end{figure}

After building a 3D map, we carry out several pre-processing steps to support the main techniques. 
As the UAV covers a larger area, the dataset quickly increases in size due to the immense amount of incoming data. It was necessary to down-sample the data to not only reduce the data size but also reduce the noise in the data. We do this using a voxel filter. The voxel filter allows down-sampling through averaging points in a set size grid. Down-sampling also helps increase the speed at which we process the data for other algorithms. 
We also apply a crop box filter to crop the point cloud data and extract only the region of interest. 
Figure~\ref{fig:wheat} shows the cropped and down-sampled map obtained after these steps. Note that while we down-sample the data here, in a later step, we will use the full dataset for a more precise estimation of the heights. 

\subsection{Plot Detection}

After the pre-processing, the next step in the pipeline is to detect individual plots within the farm. Recall, that a ``plot'' refers to a cluster of plants as shown in Figure~\ref{fig:wheat_rgb}. We need this to use the ground plane and height estimation algorithm on each plot given a point cloud file of the entire farm. We show an example dataset for this in Figure~\ref{fig:wheat}. 

\subsubsection{$K$--means Clustering}
We start with a down-sampled 3D point cloud. We first filter out points below a certain z-axis value (height). We then perform $k$--means clustering over the remaining points based on spatial distance. The k-value set for the $k$--means clustering algorithm is the number of clusters we expect, or for our case, the number of plots we are looking for. 

\begin{figure}[hbt!]
    \centering{
    \imagewidth=0.43\textwidth
    {\includegraphics[width=\linewidth]{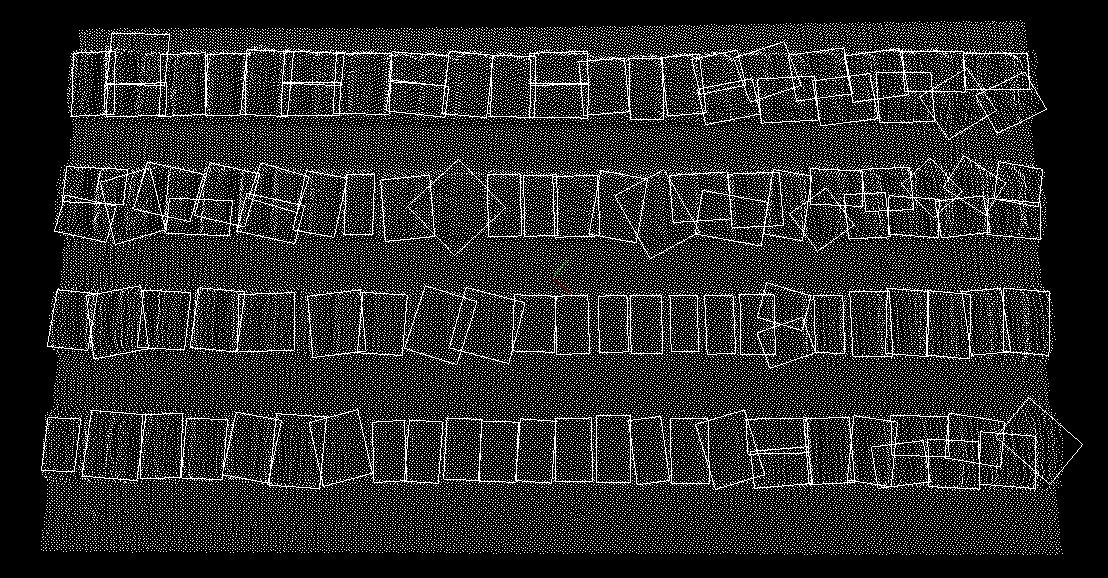}}
    \subfigure[]{}
     {\includegraphics[width=\linewidth]{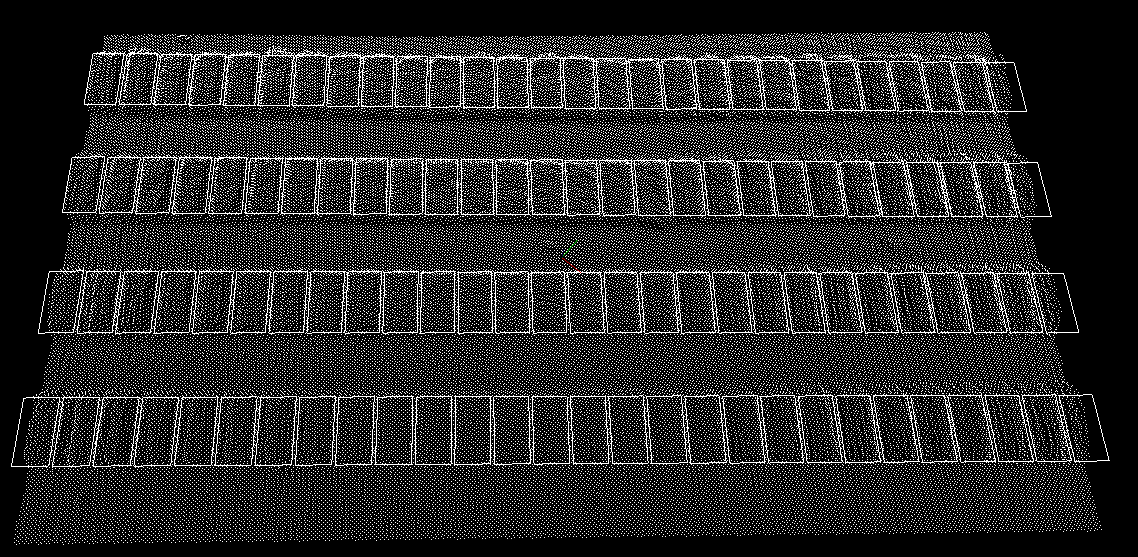}}
    \subfigure[]{}
    \caption{(a) This is example data that we fed into the $k$--means clustering algorithm. The minimum oriented boxes computed using the $k$--means clustering algorithm overlaid on top of the plots. (b) The estimated grid after the voting scheme overlaid on top of the plots. Note: While the figures were aligned with the $x$ and $y$ axes, the actual data and oriented boxes were not aligned. The algorithm also determines the rotation of the oriented boxes.}
    \label{fig:clusters}
    }
\end{figure}


In Figure~\ref{fig:clusters}.a, we show the minimum oriented bounding box of each cluster. A minimum orientated bounding box is the smallest area box that fits all the points. As shown in this figure, the algorithm does not cluster all plots correctly, but it correctly clusters a large amount of them. Because of this, we used a voting scheme to help determine the plot size using the bounding boxes.


\subsubsection{Voting Scheme for Correcting Clusters}

For the voting scheme, we use the bounding box dimensions and orientations of each of the clusters determined in the previous step. These are binned, meaning these are all broken into separate ranges. We count the number of clusters that fit into each of the ranges. After, we take the range with the largest number of votes. Then, all the values within this range are averaged and used as the orientation that we set for the plots. We use a similar method for the length and width of the clusters to determine the length and width of the plots. 
\subsubsection{Grid Selection}

Using the estimated orientation, width, and length of the plots, we fill in a grid for the dataset. The distances between the plots are manually fit to help create the grid. We highlight that even though the grid is currently manually overlaid on top of the target dataset, we are working to automate this process. After we do this grid selection, the results look similar to what we show in Figure~\ref{fig:clusters}.b. We use these bounding boxes on the original dataset. We extend the lengths of the boxes so that they cover the ground between plots. All of the points in these boxes are then extracted one box at a time and then fed into the ground plane and height estimation algorithm which we discuss next.

We assume that we know the number of plots in the farm, i.e., we know the correct number of clusters for $k$--means. However, this is not a strict requirement. We do not need to know the total number of clusters in the entire farm. During pre-processing, we can set the crop box to a small area -- small enough to manually count the number of plots. Then, we can use the $k$--means clustering followed by the voting scheme to determine the size of the cluster. Once we determine the size of the cluster and the grid pattern for the smaller cropped area, we can extrapolate that grid to the rest of the (uncropped) dataset.


\subsection{Ground Plane and Height Estimation}
The purpose of this algorithm is to take a point cloud file, find the ground plane in it, and then determine height data for objects that are not part of the ground. We describe the process of this algorithm in this section.

\subsubsection{Detect Points in Ground Plane}
We detect the ground plane in a LiDAR scan by applying a sample plane model segmentation available in PCL~\cite{Rusu_ICRA2011_PCL}. This segmentation implementation uses a random sample consensus (RANSAC) algorithm to find a plane within the given dataset. 

\subsubsection{Best-Fit Plane of Inlier Points}
We find the inlier points from the planar model segmentation and use them to find the equation that describes a best-fit plane. We do this using the linear least-squares approximation on the inlier dataset. We take each of the inlier points and use them for this approximation to output the centroid of the data, as well as the normal vector of the best-fit plane.

\subsubsection{Height Estimation}

With a centroid and normal vector for the best-fit plane, we determine the height of each of the outlier points. We do this by finding the distance of each outlier point to the best-fit plane. Since the plane describes the ground, the distance from the outlier point to the plane is the height of that point over the ground. Doing this for all of the outlier points indicates each of their heights relative to the ground plane. As stated in the grid selection subsection, we extended the bounding box lengths. We did this so that the ground plane estimation works on a localized scale relative to each plot. This helps improve the accuracy of the algorithm by adjusting for the disparities in the ground over a large field. We purposely created a virtual farm environment with ground disparities shown in Figure~\ref{fig:sim_ground}.b. The output of the local ground plane estimation is shown in Figure~\ref{fig:sim_plots}. It was later discovered that using only voxel-filtered data led to underestimations. This is to be expected since voxel-filtering averages points within a set voxel resolution. To account for this, we concatenated the raw point cloud scans of the plots with the voxel-filtered data. More details are given in Section~\ref{chap:experiments}.

\section{Farm Generation Toolchain} \label{chap:farmGen}
In this section, we describe the algorithm that generates the virtual phenotyping farm world. The virtual environment is a Gazebo simulation environment. The hector quadrotor ROS package~\cite{2012simpar_meyer} is used as the virtual UAV with a Velodyne VLP-16 model attached to the bottom. We also use commercial 3D models of individual soybean plants. These are under the standard 3D Model License of \url{https://www.turbosquid.com}. We then create an automatically generated farm model from the individual plants. The algorithm is composed of two main parts, random ground creation and plot generation, discussed below.

\subsection{Random Ground Creation}
The first step of the farm generation algorithm is to create a ground model based on the following input parameters. We take as input the size of the square farm. We also take as input a parameter that gives the number of random vertices that are generated within the square. These vertices are then used to create a Delaunay triangulation model of the environment. The last parameter is the range within which the heights of each vertex are randomly varied. This creates a ground model that has a random variation in height and is represented as a triangulated terrain. 

\subsection{Plot Generation}
\begin{figure}[hbt!]
    \centering{
    \imagewidth=0.43\textwidth
    {\includegraphics[width=\linewidth]{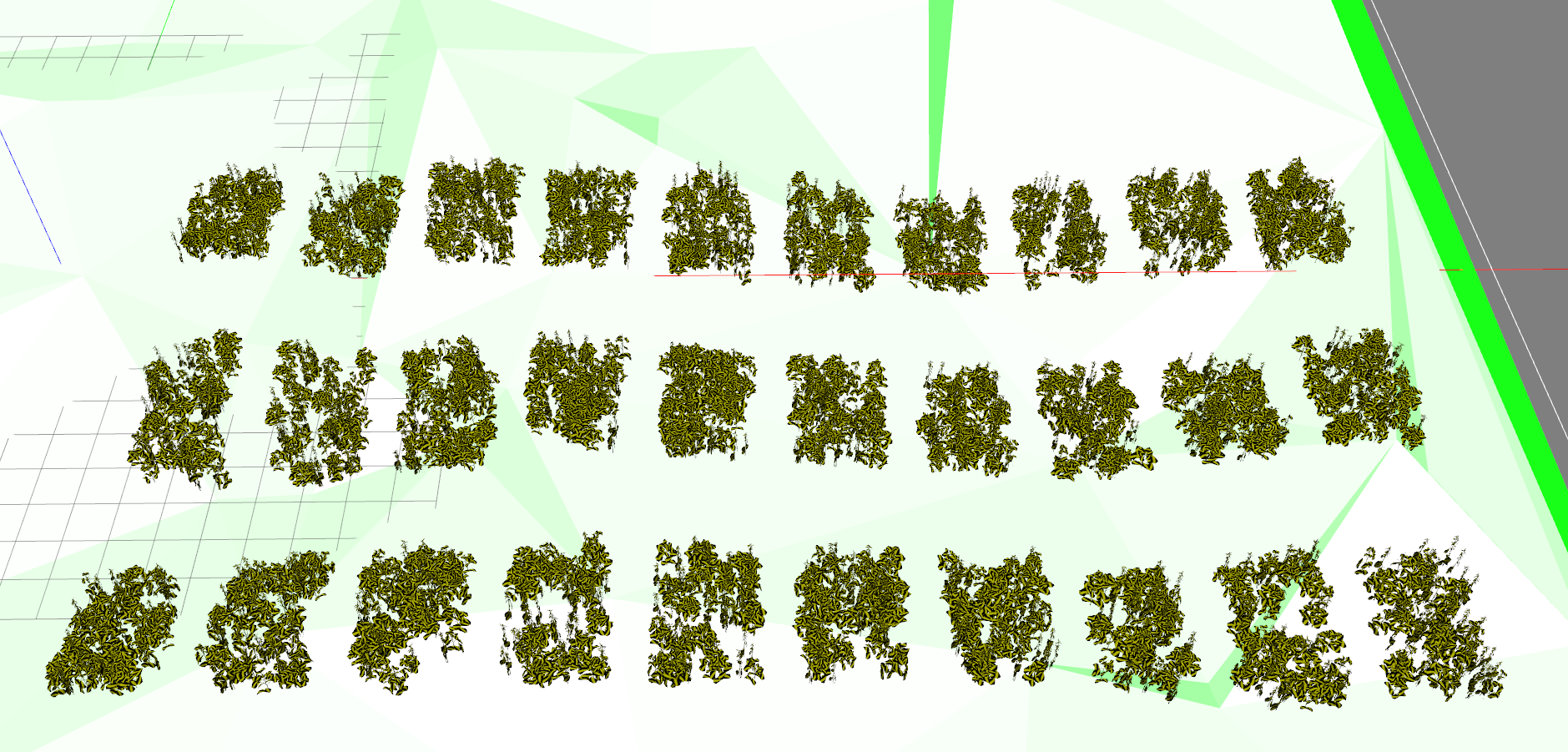}}
    \subfigure[]{}
    \caption{Screenshot taken of one of the generated phenotyping farms. This shows an overhead view of the entire farm. While it is hard to see in the figure, the ground model in this environment does vary between $\pm1$ m.}
    \label{fig:gazebo}
    }
\end{figure}

Next, we create individual plots within this virtual farm. This step takes as input the following parameters: the width and length of the plots, the width and length between plots, the number of plants within each plot, the minimum and maximum scaling applied to individual plant models, and the row data. The row data is comprised of the number of plots within rows and the offset of where the row starts. Using these parameters, the individual plant models are randomly composed into the described environment. They are randomly scaled and rotated within the minimum and maximum range. Their heights are adjusted based on the height of the terrain at that point. An example of this is shown in Figure~\ref{fig:gazebo}.

\section{Experiments, Simulations, and Results} \label{chap:experiments}

We conducted experiments and data collection at Virginia Tech's Kentland Farm. Kentland Farm provided real-life examples of plotted crops that we could use to collect data. We describe the experiments in detail in this section, along with their results. We also describe simulations we conducted using the virtual farm generation tool that was developed and provide the results of those simulations.

\subsection{Kentland Experiments}

We performed real-world testing of our algorithm at Virginia Tech's Kentland Farm on a set of wheat plots maintained by Carl Griffey, Ph.D. for state variety testing in Spring 2019. Figure~\ref{fig:wheat} shows the processed point cloud data that we worked with and is comprised of 112 wheat plots. Flights were conducted during Spring and Summer 2019. However, for the algorithm pipeline, only a single flight is needed. For the wheat dataset provided, the flight was conducted in June of 2019 during the peak height of the wheat crops.

The goal of our plot detection algorithm was to correctly find a bounding box for each of these plots. Figure~\ref{fig:clusters}.a highlights the bounding boxes when using just a $k$--means clustering algorithm to determine clusters within the point cloud data. As stated before, the algorithm correctly found many of the plots/clusters, but some did not fit at all. We fed the dimensions of the bounding boxes into the voting scheme step of the plot detection algorithm. Of the total 112 plots, the output of the voting scheme found 52 plots that fit within a certain orientation as well as width range, whereas, 54 were found that fit within a length range. We averaged all of the values within the range to determine the best-fit orientation, width, and length of the bounding boxes for the plots. The grid was then manually setup to overlay on top of the point cloud data to extract the points within each plot. We showed this grid fitting previously in Figure~\ref{fig:clusters}.b.

\begin{figure}[htp]
\centering
\includegraphics[width=\linewidth]{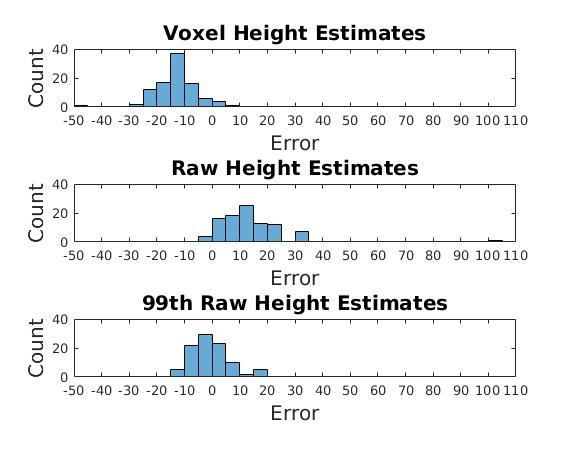}
\caption{The results of the Kentland wheat farm experiment. The top histogram shows the error between the manual measurements and the voxel-only estimations. The middle histogram shows the error using the raw data estimations. The bottom histogram shows the error using the 99th percentile of the raw data estimations.}
\label{fig:p_dif}
\end{figure}

The height estimation on the farm field gave promising results. We manually measured 3 plants within each plot of wheat. We averaged these 3 measurements to get a plot height. Figure~\ref{fig:p_dif} shows the error between the hand-measurements and the height estimation. For the height estimation, we tried a few different methods. The first method was to use the voxel filtered data set shown in Figure~\ref{fig:wheat}. We used the maximum height within each plot as the height estimate. We show this in the first histogram of Figure~\ref{fig:p_dif}. Since it was a down-sampled dataset, the majority of the output was underestimations and the average error between the manual measurements and height estimations was $\pm13.4$ \%. We show in the figure that the errors center around the -15 to -10 \% bin. The next method was to fit the raw plot data over the voxelized ground data. We cropped and recentered the raw data, but did not down-sample it using the voxel filter. Afterward, we used the plot detection bounding boxes to extract the points within each plot. These were then concatenated with the already voxel filtered dataset in Figure~\ref{fig:wheat}. The maximum height within each plot was again used. We show the results of this methodology in the second histogram in Figure~\ref{fig:p_dif}. These gave less accurate results. Because we used the raw data, the new dataset was noisier and the error was $\pm13.8$ \%. The error centered around the +10 to +15 \% bin for this method, but there was also a single data point within the +100 to +105 \% bin due to noise. To get more accurate results, we extracted the 99th percentile height point from each plot. This gave the most accurate height estimations with the error shown in the bottom histogram of Figure~\ref{fig:p_dif}. The average RMSE across all plots was 6.1 cm using this method with an error of $\pm5.4$ \%. The error for this method centered around the -5 to 0 \% bin of the figure. The noisy outlier point in the raw-only methodology was also removed using this.

\begin{figure}[htp]
\centering
\includegraphics[width=\linewidth]{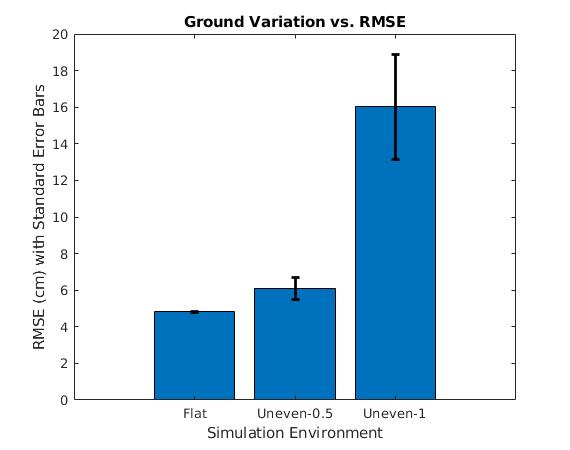}
\caption{The results of the simulated farm world experiments. The columns represent the 3 different simulation environments: one with a flat ground, one with a ground where the height varies $\pm0.5$ m, and one where the height varies $\pm1$ m. As seen in the figure, the more the height of the ground varies, the greater the RMSE and standard error is.}
\label{fig:sim_results_bar}
\end{figure}


\subsection{Soybean Plot Simulations}
\begin{figure}[hbt!] 
  \subfigure[]{%
    \includegraphics[width=.2\textwidth]{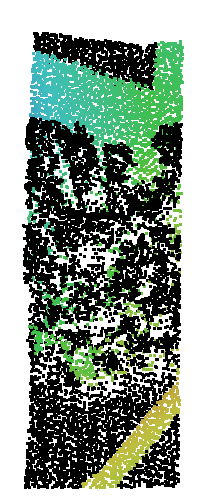} \label{fig:plot1} 
  } 
  \quad 
  \subfigure[]{%
    \includegraphics[width=.2\textwidth]{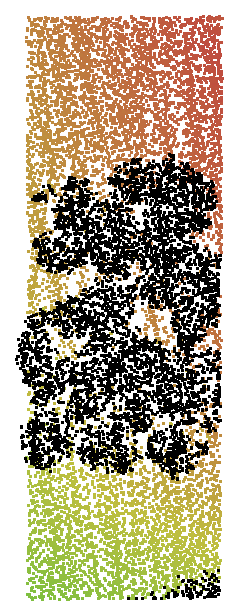} \label{fig:plot2} 
  } 
  \caption{Zoomed in plots that are shown in Figure~\ref{fig:sim_ground} to highlight the ground plane estimation algorithm. These 2 plots are placed on a ground that varies. The colored points show points that lie within the ground plane while the black points are outliers. (a) The ground in this plot varies a lot compared to other plots and it can be seen that the colored points are not all of the ground points. (b) This ground varies also, but not as much. Almost all of the ground points in this are correctly colored.} 
  \label{fig:sim_plots}
\end{figure}

\begin{figure}[hbt!]
    \centering{
    \imagewidth=0.43\textwidth
    {\includegraphics[width=\linewidth]{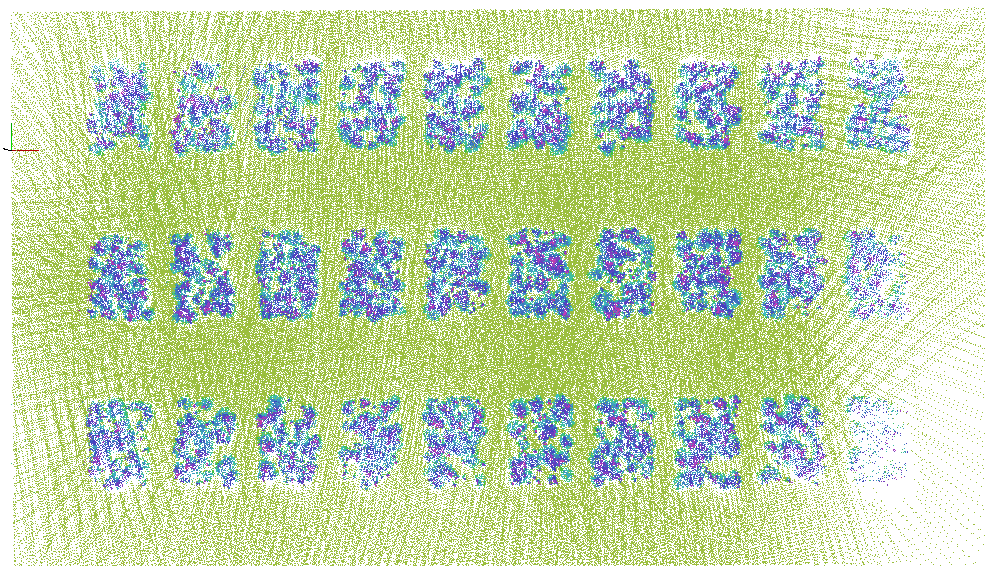}}
    \subfigure[]{}
     {\includegraphics[width=\linewidth]{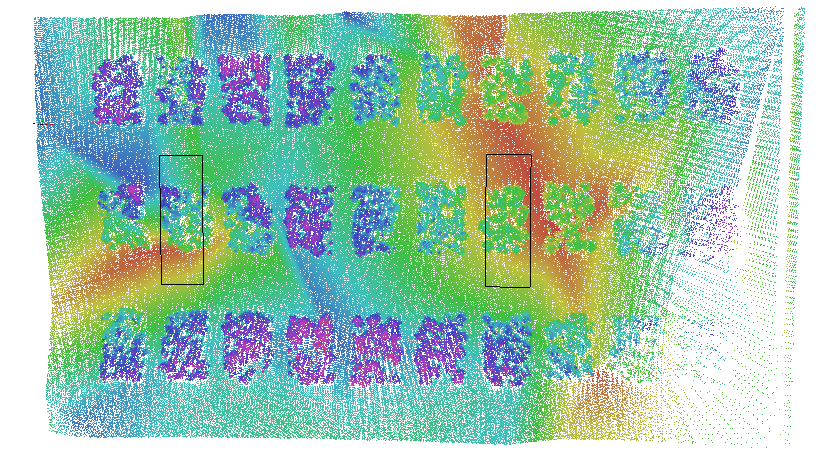}}
    \subfigure[]{}
    \caption{2 different simulation environments. The different RGB values correspond to the height (z-value) of the points. (a) A simulated soybean farm world with a flat ground model. (b) A simulated soybean farm world with a ground model with 150 vertices that have $\pm1$ m height values.}
    \label{fig:sim_ground}
    }
\end{figure}

We also performed simulations on 3 different environments generated using our farm generation toolchain. These environments consisted of the same number of plots. There were 3 rows of 10 plots, each consisting of 90 soybean plants. 
The difference between the 3 environments is the ground terrain. 2 of these environments are shown in Figure~\ref{fig:sim_ground}. 
Notably, Figure~\ref{fig:sim_plots} highlights the ground plane estimation algorithm. Figure~\ref{fig:sim_plots}.a consisted of a more extreme ground; however, the algorithm still estimates a good ground plane. Figure~\ref{fig:sim_plots}.b still had a ground model that varied, but not as intensely. This environment represents a more realistic scenario for a farming environment. The algorithm was able to find more of the ground points accurately. 

The results of the simulations are shown in Figure~\ref{fig:sim_results_bar}. As expected, the environment with the flat ground model performed the best with an RMSE of 4.8 cm. The more realistic ground model is shown in the second column and resulted in an RMSE of 6.1 cm. Lastly, the environment with the more extreme ground model is shown in the third column and had an RMSE of 16.0 cm. As expected, as the ground varies more extremely, the RMSE of the height estimation algorithm increases. Also of note, the standard error of the mean increases correspondingly.

\section{Future Work} \label{chap:conclusionFutureWork}
In this paper, we present several tools that are useful in high-throughput phenotyping of wheat and soybean plants. Specifically, we focus on easing the data processing pipeline of high-throughput phenotyping with a 3D LiDAR mounted on a UAV. We present point cloud processing techniques that produce as output a 3D model of the farm, semi-automatically find individual plots within a farm, and estimate the height of crops in each plot. In addition to hardware experiments, we also present a simulation toolchain to randomly produce virtual farms to test these algorithms. Our algorithms can estimate heights within an accuracy comparable to other methods. More importantly, our algorithms can produce other outputs (such as a 3D map and individual plots) that are needed for phenotyping that prior work has not focused on. An immediate avenue of future work is to test the algorithm for other types of farms besides the plot-based phenotyping farms considered in this paper. The data processing software, real-world datasets, and simulation toolchain are released along with this paper. We hope that this will make agriculture research more accessible to robotics research by reducing the hardware barrier to entry.

\addtolength{\textheight}{-12cm}   





\addcontentsline{toc}{section}{Bibliography}
\bibliographystyle{unsrt}
\bibliography{ref}

\end{document}